# Bidirectional Chinese and English Passive Sentences Dataset for Machine Translation

**Xinyue Ma**[1,2]**, Pol Pastells**[1,2]**, Mireia Farrús**[1,2]**, Mariona Taulé**[1,2]

[1]Centre de Llenguatge i Computacio (CLiC), Universitat de Barcelona, Spain
[2]Institut de Recerca en Sistemes Complexos (UBICS), Universitat de Barcelona, Spain
`{maxinyue, pol.pastells, mfarrus, mtaule}@ub.edu`

**Abstract**

Machine Translation (MT) evaluation has gone beyond metrics, towards more specific linguistic phenomena. Regarding English-Chinese language pairs, passive sentences are constructed and distributed differently due to language variation, thus need special attention in MT. This paper proposes a bidirectional multi-domain dataset of passive sentences, extracted from five Chinese-English parallel corpora and annotated automatically with structure labels according to human translation, and a test set with manually verified annotation. The dataset consists of 73,965 parallel sentence pairs (2,358,731 English words, 3,498,229 Chinese characters). We evaluate two state-of-the-art open-source MT systems with our dataset, and four commercial models with the test set. The results show that, unlike humans, models are more influenced by the voice of the source text rather than the general voice usage of the source language, and therefore tend to maintain the passive voice when translating a passive in either direction. However, models demonstrate some knowledge of the low frequency and predominantly negative context of Chinese passives, leading to higher voice consistency with human translators in English-to-Chinese translation than in Chinese-to-English translation. Commercial NMT models scored higher in metric evaluations, but LLMs showed a better ability to use diverse alternative translations. Datasets and annotation script will be shared upon request*.

## 1. Introduction

Although passive sentences can be found in both Chinese and English, the constructions used to express passive voice in these two languages differ in form, frequency and distribution. Passive voice is relatively common in English, conferring a sense of objectivity. It is widely used in press, official documents and academic writings, while in Chinese, its usage is far more restricted. Passive voice is approximately eight times more common in English than in Chinese (Dong et al., 2023). Chinese passives are mainly used to express unpleasant feelings and unfavourable events (Xiao et al., 2006). Therefore, due to this obvious discrepancy, special attention should be paid to passive sentences in translation tasks in order to convey equivalent information into the target text.

Human translators employ several strategies to translate passive sentences in both languages. The most straightforward is to maintain the passive voice in the target text. In English, the most common passive construction is *be + past participle*, where *be* can be replaced with auxiliaries such as *get*, *have*, *become*, or *remain*. Note that, unlike *be*, which is neutral and not sensitive to the semantic feature of dynamicity, *get* only occurs in dynamic events (Quirk, 1985) and prefers negative context (Xiao et al., 2006). In Chinese, *bei* 被 *(+ NP) + VP* is the most frequent passive structure. Apart from *bei*, there are other passive markers such as *rang* 让, *gei* 给, *jiao* 叫 and light verbs that are delexicalized and mainly serve the purpose of expressing the passive voice, such as *shou* 受 and *zao* 遭 (Cai et al., 2019). All these passive constructions in Chinese are mainly used for unfavourable events, which forces translators to turn to the active voice when translating English passives with neutral and positive content into Chinese.

If the active voice is employed, the original patient must change its role from subject to object. A generic agent such as *they* or *people* might be added as the subject if no agent is mentioned in the original passive structure. This approach applies to both Chinese and English. However, there are many other options available for English-Chinese translation (Kefei and Dingjia, 2018). If the patient is the focus, a topic sentence with a copula *shi* 是 in which the patient remains the subject can be used; in the construction *light verb (LV) + verbal noun (VN)*, the verbal noun conveys the main action in the passive structure (similar to the English LV+VN structure in "take a look"); resultatives marked with *ba* 把 and *jiang* 将 or causatives marked with *shi* 使 and *ling* 令 can also express

---

the idea that the patient undergoes a certain action. Trained human translators have sufficient experience and knowledge to choose the appropriate translation strategy and modify the syntactic structure accordingly, but the capabilities of MT models on this task remain to be evaluated.

Previous studies of Chinese and English passives investigate either their linguistic features as original text (Xiao et al., 2006) or focus on only one direction of human translation (Dai and Xiao, 2011; Kefei and Dingjia, 2018). Bidirectional research of machine translation performance has been conducted for English-French passives (Henkel, 2024) but not between Germanic languages and Sino-Tibetan languages. With the purpose of filling this gap, in this paper, we present an annotated bidirectional multi-domain dataset of Chinese and English passive sentences, which can be used to study human translation strategies, as well as to evaluate MT model performance. Moreover, the dataset can also serve the purpose of fine-tuning MT models for better performance on passive sentence translation.

The main contributions of this paper are summarised as follows:

1. We propose a bidirectional multi-domain parallel dataset of 16,850 Chinese *bei* passives and 57,115 English *be* passives, annotated automatically for translation strategies according to dependency information. We also present a test set with manually validated annotations.

2. We evaluate six state-of-the-art MT models with our dataset and test set, and show that evaluation metrics cannot always reflect the ability of the models to make human-like choices in voice and use diverse alternative translations for passive sentence translation.

3. Our bidirectional parallel dataset also serves the purpose of performing a contrastive analysis of Chinese and English passive structures, and of passives in original and translational texts.

## 2. Related Work

Passive sentence translation for MT models is a topic of interest for many language pairs. Kim (2022) explored how Korean passives are translated into English. The difficulty in translating passives lies in the fact that Korean has three different passive constructions and various affixes that play an important role in expressing passives. The author tested Google Translate (GT) and Naver Papago (NP) on a dataset of 213 passives, and the commercial models performed very well at capturing the correct meaning, and they sometimes output the English translation in the active voice (22.4% for GT and 31.63% for NP). Henkel (2024) studied English (*be* passive) and French (*être* + past participle) passives in human and machine translation using a self-constructed bidirectional corpus (±13.5 million words). The results showed that, compared to the original French text, the English to French translation by DeepL Pro contained approximately 40% more passive constructions, while human translation showed no significant discrepancy. Also, human translators of both languages showed greater diversity in adopting alternative translation counterparts. Chang et al. (2013) focused on Chinese to English patent document machine translations and proposed a set of transformation rules for Chinese marked passives, notional passives, and other structures so that they are translated correctly into English passives.

As for MT model evaluation, there are studies covering a variety of linguistic features. Guillou et al. (2018) tested 16 English-to-German MT models on pronoun translation. They found that NMT models perform well if there is intra-sentential reference, but the inter-sentential cases remain difficult. Macketanz et al. (2018) constructed a test suite of 5000 sentences covering 106 linguistic phenomena, including passive voice, to perform a fine-grained evaluation of 16 German-to-English MT models. Punctuation, multi-word expressions, ambiguity, false friends and future tense are found to be generally difficult for models. Popović (2019) proposed an evaluation dataset for MT performance on ambiguous conjunctions from English (*but*) and French (*mais*) to German (*aber* or *sondern*). The results showed that multilingual training is beneficial for conjunction disambiguation. Song and Xu (2024) focused on multi-word expressions (MWEs) in Chinese-to-English. Their study showed that MWEs are difficult for MT models to translate, and MT evaluation metrics tend to overrate the translation of sentences containing MWEs. Song et al. (2024) has a wider scope and created a multi-domain test suite annotated with 43 grammatical features for Chinese-to-English MT evaluation. However, the scores were produced by metrics, and no further investigation was conducted to analyse the translation strategy adopted by MT models.

## 3. Dataset Construction

In order to examine and evaluate the translation strategies adopted by MT models compared to human translation, we constructed a multi-domain bidirectional parallel dataset of Chinese and English passive sentences. Data is extracted for the two most common passive constructions, namely *bei* passive and *be* passive, from five parallel cor-pora. All their occurrences in source and target texts in all possible genres are included, yielding a dataset that consists of 73,965 parallel sentence pairs (2,358,731 English words and 3,498,229 Chinese characters).

### 3.1. Data Source

We collected parallel data of Chinese and English passive structures from five publicly available corpora, all aligned at the sentence level.

1. **CECPC-Core.** China English-Chinese Parallel Corpus-Core is a bidirectional corpus including literary and non-literary texts in technology, popular science, social sciences, News, and many other areas. It contains around 18 million Chinese characters/words, and all data is tokenized and PoS-tagged (tagsets unspecified in documentation).

2. **Yiyan.** (Xu and Xu, 2021) The Yiyan English-Chinese Parallel Corpus contains 2.6 million Chinese characters/words (1 million words of English texts translated into 1.6 million Chinese characters). Tokenization and PoS-tagging were realized with spaCy ("zh_core_we_sm" for Chinese texts and "en_core_web_sm" for English texts).

3. **BABEL.** (Xiao, 2004) The Babel English-Chinese Parallel Corpus consists of 327 English articles from two magazines and their translations in Chinese. It contains 544,095 tokens (253,633 English words and 287,462 Chinese characters) in total. The English texts were tagged using the CLAWS C7 tagset while Chinese texts were tagged using the Peking University tagset.

    These three corpora are available at CQPweb (Hardie, 2012), a web-based corpus analysis system created by Andrew Hardie (2012).

4. **E-C Concord.** (Wang, 2001) English-Chinese Parallel Concordancer is a bidirectional English-Chinese parallel corpus which includes literature, legal documents and academic writing. It contains 4.5 million Chinese characters and 2.9 million English words. No annotation was added.

5. **Chinese-English Parallel Corpus of Classic Chinese Literature** (中国经典文学作品汉英平行语料库, hereinafter ClassicCL). (Feng et al., 2018). This corpus consists of 50 Classic Chinese literary works and one or more versions of their English translation. Texts are tokenized and PoS-tagged using the CLAWS7 Tagset for English and ICTPOS3.0 for Chinese. English texts are also annotated with the UCREL Semantic Analysis System (USAS) for semantic tags.

### 3.2. Register Categorization

Since each corpus covers different genres and has different criteria for categorization, we later checked each genre manually and recategorized the data according to personal judgment into five registers: press, legal document, academic prose, general prose, and literature, as shown in Table 1. The original genre or specific book titles are kept as additional information if available.

### 3.3. Passive Data Extraction and Cleaning

We searched for *bei* and *be* passives in both source texts and target texts. For the three CQPweb corpora, we searched for the *bei* token PoS tagged as a preposition for *bei* passives, and the lemma *be* plus a past participle within four words for *be* passives. For E-C Concord and ClassicCL, we can only search by word form. For Chinese, we searched the word *bei* and for English, all the inflected forms of *be*. Information about the genre and translation direction of each match is also retrieved and kept.

In order to rule out non-passives and adjectival passives (such as "I am worried"), we used spaCy to check the dependency label of the lemma *be* in all the *be* passives. A match was kept if the lemma *be* was labelled "auxpass". We filtered Chinese *bei* passives with the PoS tag. Matches in which *bei* was tagged as "LB" or "SB" (meaning *bei* passive with or without agent) were kept.

We also dropped cases that were excessively long (the English part exceeded 100 words), which was often caused by a poor paragraph truncation. Cases in which the length of the Chinese and English parts were disproportionately uneven were also deleted. They were either the result of over-explicitation or simplification, or were caused by an incorrect sentence alignment in some corpora. The ratio we used for Chinese characters/English words was 0.5–2.2, which was quite loose in order to keep most valid cases.

After cleaning the dataset, we have 16,850 sentence pairs for *bei* passives and 57,115 for *be* passives, which is consistent with the fact that the passive voice is much more frequent in English than in Chinese. We then separated these pairs into four subsets and named them according to the translation direction and their occurrence in the source or target texts. **ZH→EN** stands for Chinese-to-English translation, that is, only contents originally written in Chinese and translated into English qualify, and **EN→ZH** for English-to-Chinese translation. Note that there is some overlap among the subsets.

| Register | Genre | | | | |
|---|---|---|---|---|---|
| | CECPC-Core | Yiyan | BABEL | E-C Concord | ClassicCL |
| A_PRESS | A_Reportage<br>F_Arts, Trade, Entertainment | A_Reportage<br>B_Editorials<br>C_Reviews<br>H_Reports, Gov_Docs | / | / | / |
| B_OFFICIAL DOCUMENT | D_Government Documents<br>E_Legal Documents | / | / | Legal Documents | / |
| C_ACADEMIC PROSE | H_Miscellaneous<br>I_Technical Text | J_Academic Writing | / | Academic Writing | / |
| D_GENERAL PROSE | G_Philosophy, Social Sciences<br>J_Popular Science Text | D_Religion<br>E_Skills, Hobbies<br>F_Popular Lore<br>G_Biography, Memoirs | Magazine | Speech | / |
| E_LITERATURE | K_Pure Literature, Biography, Essay<br>L_General Fiction<br>M_Mystery and Detective Fiction<br>N_Science Fiction<br>P_Drama Legend<br>R_Children Literature | K_General Fiction<br>L_Mystery Detective<br>M_Science Fiction<br>N_Adventure Fiction<br>P_Romance Stories<br>R_Humor | / | Novel<br>Essay | Literature |

Table 1: Register categorization. Decision was made after checking the content of each genre. Some genres with the same name contain very different texts in different corpora.

| Direction | Register | Corpora | | | | | Sum |
|---|---|---|---|---|---|---|---|
| | | CECPC-Core | Yiyan | BABEL | E-C Concord | ClassicCL | |
| ZH(bei)→EN | A | 230 | / | / | 3 | / | 233 |
| | B | 370 | / | / | / | / | 370 |
| | C | / | / | / | / | / | 0 |
| | D | 167 | / | / | / | / | 167 |
| | E | / | / | / | 2,809 | 5,765 | 8,574 |
| | **Sum** | **767** | **0** | **0** | **2,812** | **5,765** | **9,344** |
| EN→ZH(bei) | A | 213 | 640 | / | / | / | 853 |
| | B | 999 | / | / | 48 | / | 1,047 |
| | C | 87 | 412 | / | 31 | / | 530 |
| | D | 1,493 | 887 | 479 | 2 | / | 2,861 |
| | E | 11 | 514 | / | 1,690 | / | 2,215 |
| | **Sum** | **2,803** | **2,453** | **479** | **1,771** | **0** | **7,506** |
| **Total** | | **3,570** | **2,453** | **479** | **4,583** | **5,765** | **16,850** |

Table 2: Distribution of *bei* passives.

| Direction | Register | Corpora | | | | | Sum |
|---|---|---|---|---|---|---|---|
| | | CECPC-Core | Yiyan | BABEL | E-C Concord | Classics | |
| ZH→EN(be) | A | 3,484 | 49 | / | / | / | 3,533 |
| | B | 5,361 | / | / | / | / | 5,361 |
| | C | / | / | / | / | / | 0 |
| | D | 2,721 | / | / | / | / | 2,721 |
| | E | / | 7,151 | / | / | 13,197 | 20,348 |
| | **Sum** | **11,566** | **7,200** | **0** | **0** | **13,197** | **31,963** |
| EN(be)→ZH | A | / | / | 1,920 | / | / | 1,920 |
| | B | 80 | 130 | / | / | / | 210 |
| | C | 364 | 113 | 1,853 | / | / | 2,330 |
| | D | 7,449 | 7 | 2,366 | 1405 | / | 11,227 |
| | E | 2,654 | 5,652 | 1,159 | / | / | 9,465 |
| | **Sum** | **10,547** | **5,902** | **7,298** | **1,405** | **0** | **25,152** |
| **Total** | | **22,113** | **13,102** | **7,298** | **1,405** | **13,197** | **57,115** |

Table 3: Distribution of *be* passives.

If a Chinese-to-English sentence pair contains *bei* passive in the Chinese part and *be* passive in the English part, it would appear in both *ZH(bei)→EN* and *ZH→EN(be)* subsets. This way, if one intends to focus on a single passive structure, the two *be* subsets or the two *bei* subsets would contain all relevant cases. Tables 2 and 3 show the distribution of data across these four subsets ("/" indicates that no data for this register are available in the corresponding corpus).

### 3.4. Annotation of Translation Strategy

The annotation of the translation strategy is performed automatically according to the results of PoS tagging, syntactic parsing and semantic dependency parsing by spaCy[1] (Honnibal et al., 2020) and LTP[2] (Che et al., 2021). Based on the reported accuracy, we chose the "en_core_web_trf" spaCy model for the English PoS tagging and syntactic dependency parsing tasks. For Chinese, we used "zh_core_web_trf" for PoS tagging and the LTP/base2 model for syntactic dependency parsing and semantic dependency parsing.

As mentioned briefly in the introduction, many different structures can be used to translate passives (or be translated into passives) in Chinese and English. The situation of the corresponding Chinese of *be* passives is complex. We set rules to capture 1) *bei* 被 passives, 2) other marked syntactic passives, 3) lexical passives, 4) notional passives, 5) *ba* 把 and *jiang* 将 resultatives, 6) patient-subject sentences with *you* 由, 7) patient-subject sentences with the copula *shi* 是, 8) light verb + verbal noun structures, and 9) *shi* 使 and *ling* 令 causatives. The remaining structures are labelled with N/A and are assumed to be sentences in the active voice.

---

[1] Glossary of all spaCy parsers available at: https://github.com/explosion/spaCy/blob/master/spacy/glossary.py

[2] Glossary of all LTP parsers available at: https://ltp.ai/docs/appendix.html#

The annotation for English is relatively easy, since passive markers can be identified accurately by models. For the 16,850 sentence pairs that contain *bei* passive in Chinese, we set three rules to categorize the corresponding English sentences, yielding four passive markers: *be*, *get*, *become* and *have*. The remaining structures are labelled with N/A and assumed sentences in active voice. Notional passives are not distinguished from active sentences in English, since current semantic parsers for English parse the patient in a notional passive as "ARG1", the same as the agent in an active sentence. All the labels for translation strategy and the corresponding rules are listed in Table 4.

However, not all cases are labelled correctly due to dependency parsing errors and construction diversity. Also, the occurrence of the listed structures might not correspond to the translation of the passive structures captured during data collection, but to the translation of other parts. In order to analyse specific cases in greater detail and acquire reliable evaluation results, we manually built a test set to validate the translation strategy of 50 sentence pairs for each register that has data for *be* and *bei* passives in the source text. This yielded 200 cases from the *ZH(bei)→EN* subset (we have no data for academic prose in this direction) and 250 cases from the *EN(be)→ZH* subset. We made sure that the cases in each register reflect the diversity in translation strategies, and that the proportion of data sources (i.e. from which corpus) and the frequency of each structure were adjusted according to the composition of the whole dataset. The test set is small due to limited manpower, but its smaller size allows us to test it on commercial models for free.

## 4. Experiments and Result analysis

### 4.1. Experimental set up

We tested two state-of-the-art open-source NMT models on our dataset with four subsets, namely *Helsinki-NLP/opus-mt-en-zh* (zh-en) and *facebook/nllb-200-3.3B*. All four subsets were translated in line with the original translation direction, and annotated for translation strategy according to the rules mentioned above. We then compared the annotation of model translations with that of the original dataset.

We translated the small test set we validated with the following commercial models: Google Translate, DeepL Classic model, DeepL Next-gen model and GPT-5. The results of the two open-source models are also compared here. The translation strategies used are first annotated automatically and then corrected by hand.

To evaluate their performance, we first applied three automatic evaluation metrics that use a reference file: BLEU (Papineni et al., 2002), chrF++[3] (Popović, 2017) and COMET (Rei et al., 2022). Then we evaluated the models according to our annotation, mainly considering whether the models adopted the same voice and translation strategy as human translators did.

We apply different criteria for Chinese and English. For English, each label is considered a category, yielding five in total. Sentences labelled with *be*, *become*, *get*, and *have* are in the passive voice, while those labelled with N/A are considered to be in the active voice. For Chinese, since the possible structures that can be used to translate a passive sentence are diverse, we categorized the labels into eight translation strategies according to syntactic structure: 1) syntactic passive, 2) lexical passive, 3) notional passive, 4) topic sentence, 5) light verb structure, 6) causative, 7) resultative and 8) other active sentence. Only the first two kinds qualify as the passive voice, and the remaining six are considered active sentences. The notional passive is deemed to have the "middle voice" by some research (see Ting (2006)), but it can also be considered a topic sentence that can express the passive meaning because of the nature of its subject (Xiao et al., 2006), see (1) as an example. We do not attempt to tackle this problem here and categorize notional passives as the active voice, along with other topic sentences. In total, 18 different labels are given to the human translations of the *EN(be)→ZH* test set during manual validation.

(1) 饭　　烧-好-了
    fan   shao-hao-le
    meal  cook-ready-ASP
    'The meal is ready'

### 4.2. Evaluation on Dataset

We first annotated all four subsets and calculated the proportion of structures in the corresponding source texts and target texts of *bei* and *be* passives. The results are shown in Table 5 and 6.

Bidirectional data reveals the difference between original texts and translational texts. From Table 5, we can see that nearly 90% of *be* passives are translated from Chinese active sentences, acknowledging the great contrast between the fre-quency of Chinese and English passives. Comparing the source text of *ZH→EN(be)* subset and the target text of *EN(be)→ZH* subset, passive voice is more frequent in the Chinese translation of *be* passives than in the original Chinese, showing a tendency of source language shining through, which is in line with previous study (see (Dai and Xiao, 2011)). Table 6 shows that around two-thirds of *bei* passives are translated from English passives, yet when a *bei* passive occurs in the source text, its translation is more likely to be an English active sentence, suggesting source language interference.

---
[3]BLEU and chrF++ are computed using SacreBLEU (Post, 2018)

| Language | Translation strategy | | Label | Rule |
|---|---|---|---|---|
| Chinese | Marked passive | Syntactic passive | L 被 *bei* (with agent), S 被 *bei* (without agent), 给 *gei*, 让 *rang*, 为 *wei* | passive marker | 1. For bei passive: 被 *bei* appears in the tokens and PoS tagged as "LB" or "SB" by spaCy. 2. For other passive markers: Parse for syntactic and semantic dependency with LTP. Check if object appear before verb (fronting-object, FOB), and the patient (PAT), the passive marker (mRELA, marker of relation), the possible agent (AGT) and verb (ROOT) appear in this order: PAT → mRELA → (AGT) → ROOT. The head of PAT should be ROOT. |
| | | Lexical passive | 受(到) *shou(dao)*, (惨)遭(到) *(can)zao(dao)*, 挨 *ai*, 蒙 *meng* | "receive or suffer" | Label appears in the tokens. |
| | Patient-subject sentence | Notional passive | NOTIONAL | | FOB, and PAT → ROOT. No AGT should appear. No mRELA should appear between PAT and ROOT. The head of PAT should be ROOT. |
| | | Topic sentence | 由 *you* | "by" | FOB, and ordered as FOB → 由 *you*-mRELA → (AGT) → ROOT. The head of FOB should be ROOT. |
| | | | 是…的 *shi…de* | copula, "be" | 是 *shi* appears in the tokens and PoS tagged as "VC" by spaCy; 的 *de* appears after it and right before a comma or period. |
| | Light verb + verbal noun | | 得到 *dedao*, 获(得) *huo(de)*, 得以 *deyi*, 经(过) *jing(guo)*, 予(以) *yu(yi)*, 给予 *jiyu*, 加以 *jiayi*, 进行 *jinxing*, 实施 *shishi*, 付诸 *fuzhu* | "get or receive" "able to" "undergo" "give" "inflict" "conduct" "put into practice" | Label Appears in the tokens and a verb appears in the next 4 tokens without a punctuation in between. |
| | Causative | | 使 *shi*, 令 *ling* | "cause", "order" | Label appears in the tokens. |
| | Resultative | | 将 *jiang*, 把 *ba* | resultative marker | 1. For ba resultative: 把 *ba* appear in the tokens and PoS tagged as "BA" by spaCy. 2. For jiang resultative: 将 *jiang* appear in the tokens and ordered as (AGT) → 将 *jiang*-mRELA → PAT → ROOT. The head of PAT should be ROOT. |
| | Other translation in active voice | | N/A | | The rest unlabelled sentences. |
| English | Passive sentence | | BE, BECOME GET HAVE | | 1. For all passive markers: Parse with spaCy. 1) Label appears in the tokens and labelled "auxpass". 2) Label appears in the tokens and labelled "aux", the subject of the sentence is labelled "nsubjpass". 2. For get passive: *get* appears in the tokens and a past participle labelled "ccomp" appears in the next 4 tokens with get as its head. 3. For have passive: *have* appears in the tokens and a past participle labelled "ccomp" with *have* as its head appears in the next 5 tokens. At least 1 token appears between *have* and the past participle. |
| | Active sentence | | N/A | | The rest unlabelled sentences. |

Table 4: Labels for translation strategy and corresponding rules.

To evaluate the performance of OPUS-MT and NLLB models, the same proportions are calculated for their translational outputs and compared to human translation. For *EN(be)→ZH* subset, models showed a greater tendency to keep the passive voice of *be* passives in the Chinese translation. When using alternative translations in active voice, the proportions of topic sentence, causative and resultative structures in model translations are much lower than those in human translation, but the proportions of notional passives and light verb structures do not vary a lot. This may suggest that these models (especially NLLB) may have difficulty in paraphrasing an action into a statement, and in doing a subject-object switch to use active voice, which is not required in a notional passive or a light verb structure.

| Voice | Structure | ZH→EN(be) source text | EN(be)→ZH target text | | EN→ZH(bei) target text | |
|---|---|---|---|---|---|---|
| | | | Human | OPUS | NLLB | OPUS | NLLB |
| Passive | Syntactic passive | 8.9 | 13.8 | 26.7 | 26.5 | 43.2 | 40.3 |
| | Lexical passive | 2.3 | 3.3 | 3.7 | 1.9 | 3.0 | 2.3 |
| | Notional passive | 0.9 | 0.6 | 0.4 | 0.4 | 0.4 | 0.3 |
| Active | Topic sentence | 4.7 | 6.0 | 2.9 | 0.3 | 1.2 | 0.2 |
| | Light verb | 5.2 | 3.3 | 4.2 | 2.8 | 3.3 | 2.5 |
| | Causative | 1.4 | 3.4 | 2.2 | 1.3 | 1.9 | 1.0 |
| | Resultative | 5.3 | 6.6 | 3.0 | 2.2 | 3.2 | 2.3 |
| | N/A | 71.5 | 63.1 | 57.0 | 64.7 | 43.8 | 51.2 |

Table 5: Proportion (in %) of structures in Chinese source and target text

| Voice | Structure | EN→ZH(bei) source text | ZH(bei)→EN target text | | ZH→EN(be) target text | |
|---|---|---|---|---|---|---|
| | | | Human | OPUS | NLLB | OPUS | NLLB |
| Passive | BE | 65.0 | 42.4 | 74.2 | 73.8 | 44.1 | 44.8 |
| | GET | 0.9 | 0.7 | 1.0 | 0.2 | 0.3 | 0.2 |
| | HAVE | 0.2 | 0.4 | 0.1 | 0.1 | 0.0 | 0.1 |
| | BECOME | 0.3 | 0.0 | 0.0 | 0.0 | 0.0 | 0.0 |
| Active | N/A | 33.6 | 56.5 | 24.8 | 25.8 | 55.6 | 54.9 |

Table 6: Proportion (in %) of structures in English source and target text

The proportion of passives in the model translations of *EN→ZH(bei)* subset is higher than that of *EN(be)→ZH*, although a third of the source texts are in the active voice. This shows that models do have knowledge about the mainly negative context of Chinese passives and can translate an active sentence into a passive one, although they do not yet reach human-level judgement.

When translating Chinese *bei* passive into English (*ZH(bei)→EN* subset), models are using passive voice excessively, that is, 1.5 times more than human translators do. The proportion is even higher than that in *EN→ZH(bei)* source text, which shows that 66.4% of *bei* passives are translated from English passive sentences. But when translating *ZH→EN(be)* subset, the proportion of translational passives is much lower, probably due to the low frequency of passive voice in the source text.

To sum up, MT models seem to be more influenced by the linguistic features of the source text itself, rather than the linguistic characteristics of the source language, as human translators are.

### 4.3. Evaluation on Test Set

We tested two open-source NMT models, two commercial NMT models and two commercial LLMs on our bidirectional test set of *bei* and *be* passives. The performance of models is shown in Table 7. The SacreBLEU uses a Chinese tokenizer that simply separates every character with a space for BLEU metric, and the chrF++ metric can only handle Chinese text tokenized at word-level, so we adopted Jieba[4] for preprocessing before metric evaluations.

According to the evaluation of metrics, on our task of translating Chinese *bei* passives into English and English *be* passives into Chinese, commercial NMT models outperformed commercial LLMs in both directions, while the open-source models are inferior to them all. Google Translate reaches the highest score for all three metrics. OPUS-MT offers different models for each translation direction, and scores lower than NLLB, a multilingual model, in Chinese-to-English translation, but higher in English-to-Chinese translation, indicating the possibility that translating from other languages into Chinese requires targeted training.

We annotated and validated the translation strategies of all six models, and found that models vary in their ability to use alternative ways to translate passives, and that this ability cannot always be reflected by metric evaluation. On *ZH(bei)→EN* test set, NLLB used only *be* passive and active sentences in translation, while DeepL-Classic model and GPT-5 can use *become*, *get* and *have* passives appropriately according to context. As for *EN(be)→ZH* test set, most models have a similar grasp of different strategies that can be taken to translate passive sentences into Chinese, but most models show less diversity in regard to alternative translation counterparts compared to human translation, except for DeepL-Next gen model, which gave 19 different constructions in the output.

We also compared the voice and the structure adopted by model translations to those of human translations, and calculated their consistency rate for each register and the whole test set. Results are shown in Figure 1. Open-source models are still inferior, while commercial NMT models and LLMs performed equally well.

On *ZH(bei)→EN* test set, models reached the greatest consistency in Press, with GPT-5 scoring the highest, and the lowest in Literature. The greatest discrepancy between the consistency rate of voice and structure is also found in Literature. This implies that human translators use more diverse alternative translations for passives in literary texts, while models are still no equals to them on this register. DeepL-Next gen model stands out from others in literature translation, showing an ability to translate literary texts using rich alternatives and make human-like choices, adopting the appropriate structure.

On *EN(be)→ZH* test set, models reached the highest voice consistency rate on official documents and showed similar performance on the rest of the registers. The discrepancy between voice and structure consistency rate is generally greater than that of the *ZH(bei)→EN* test set. This is not unexpected due to the rich alternatives available in Chinese. Google Translate showed the highest structure consistency in official documents. Literature translation does not seem to bring extra difficulty in English-to-Chinese translation, except for the two open-source models. DeepL-Classic model scored the highest in both voice and structure consistency rate for this register.

---
[4]https://github.com/fxsjy/jieba

| Evaluation | | NMT | | | | LLM | |
|---|---|---|---|---|---|---|---|
| | | OPUS | NLLB | Google | DeepL-C | DeepL-N | GPT-5 |
| ZH(bei)→EN | BLEU | 15 | 16.9 | **22.9** | 19.6 | 19 | 15.5 |
| | chrF++ | 43.3 | 44.2 | **51.4** | 49.4 | 48.2 | 44 |
| | COMET | 74.4 | 75.4 | **80.2** | 78.3 | 79.2 | 79.1 |
| | Label div. | 3 / 5 | 2 / 5 | 4 / 5 | **5 / 5** | 4 / 5 | **5 / 5** |
| EN(be)→ZH | BLEU | 21 | 17.8 | **29.7** | 28.7 | 23.7 | 23 |
| | chrF++ | 21.4 | 18.2 | **28.7** | 28.1 | 23.7 | 23.2 |
| | COMET | 78.1 | 74.1 | **85.2** | 84.2 | 84.6 | 84 |
| | Strategy div. | **8 / 8** | 7 / 8 | **8 / 8** | 7 / 8 | **8 / 8** | **8 / 8** |
| | Label div. | 17 / 18 | 10 / 18 | 16 / 18 | 13 / 18 | **19 / 18** | 16 / 18 |

Table 7: Evaluation results using metrics and according to annotation. *div.* stands for *diversity* in *Label div.* and *Strategy div.*

Generally speaking, models reach a higher voice consistency rate when translating into Chinese, while they reach a higher structure consistency rate when translating into English. This shows that models do have knowledge about the low frequency and mainly negative context of Chinese passives. This information helps models make human-like choices for voice in translation. However, there are many alternative structures in Chinese, leading to relatively low structure consistency. When translating Chinese passives into English, models do not have such restrictions that help them to predict human choice, leading to a lower consistency rate in voice.

### 4.4. Case study

In example 2 from *EN(be)→ZH* test set, the source text is making a neutral description, and human translation used *ba* resultative structure, while DeepL-Next gen model used *bei* passive in translation. The agent here is unknown, which would be introduced with a by-phrase after the VP if it existed. When translated into a *ba* resultative structure, this unknown agent becomes subject and moves to the position before the marker *ba*. Although this case does not involve a concrete agent, this movement in logical form still exists and can cause difficulty for MT models.

(2) Now let the gas *be* rapidly *compressed* back to its initial volume, Vi.

   a. Human:
   接下来，又 把 气体 快速
   next also BA gas rapidly
   压缩 回 初始的 体积 Vi。
   compress back initial volume Vi

   b. DeepL-N:
   现在， 让 气体 被 迅速 压缩
   now let gas BEI rapidly compress
   回 初始 体积 Vi。
   back initial volume Vi

Example 3 is a case from *ZH(bei)→EN* test set and shows the subtle difference between *be* and *get* passive. The source text is a *bei* passive involving the action *dadao* 打倒 ("beat-down"), which is both negative and dynamic. Unlike the neutral *be* passive used by GPT-5, the *get* passive adopted in human translation perfectly conveys the unfavourable attitude implied by Chinese passive, and highlights the dynamicity and strength of the action "strike down".

(3) 难道不是适得其反，想要称霸的帝国主义却得到了被 打倒的结果吗？

   a. Human:
   Didn't the results turn out to be just the opposite of what they wanted? Didn't the imperialists who aimed at domination *get struck down* themselves?

   b. GPT-5:
   Isn't it ironic that the imperialists who sought hegemony ended up *being overthrown*?

In both cases, models show sufficient ability to convey the literal meaning of the source text, yet still need improvement in selecting the appropriate and informative structure according to context, to seek both semantic and pragmatic equivalence in translation tasks.

## 5. Conclusion and Future Work

This paper presents a bidirectional multi-domain dataset of Chinese *bei* and English *be* passive sentences, with rule-based automatic annotation for translation strategy, consisting of 73,965 parallel sentence pairs. Our dataset serves the purpose of observing the different characteristics of passive sentences in Chinese and English, and in original and translational text, as well as evaluating the performance of MT models in Chinese and English passive structure translation.

We evaluated OPUS-MT and NLLB models with our dataset. Results suggest that, unlike human translation, in which the source language shines through, models are more influenced directly by the syntax of the source text and are ignorant of the general voice usage of the language. In both directions, model translations are more likely to maintain the passive voice if it is found in the source text.

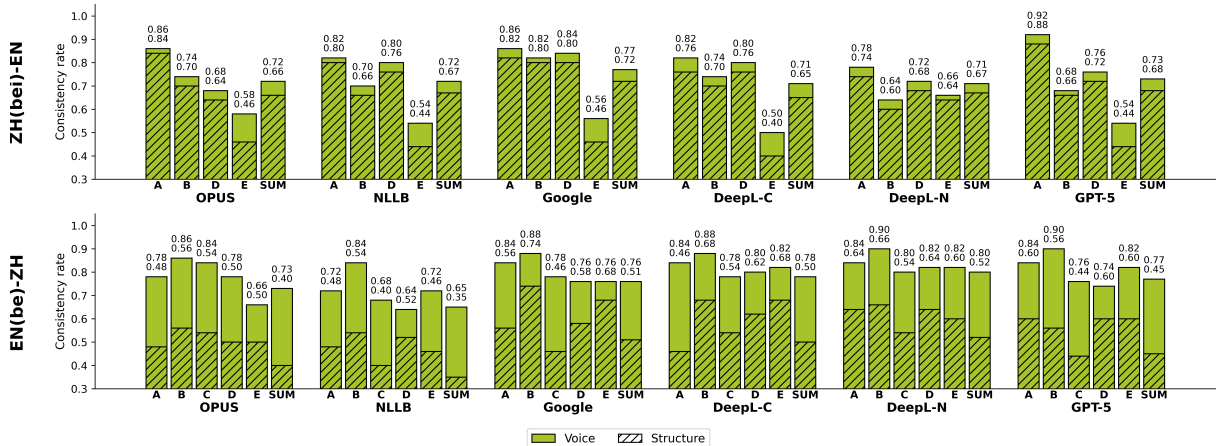

Figure 1: Consistency rate of model translations compared to human translations in the voice and the structure used, of each register and the whole test set. For translation to English, structures are categorized according to labels, while for translation to Chinese, labels belonging to the same translation strategy are regarded as the same structure, given their large number.

We conducted a more fine-grained evaluation for 4 commercial models with a manually validated bidirectional test set, and applied BLEU, chrF++, and COMET metrics to all model translations. Commercial NMT models scored higher in metric evaluations, but LLMs showed better ability in using diverse alternative structures in translations. In Chinese-to-English translation, literary text is the most difficult register for models to choose the same voice and structure as human translators do. On the other hand, voice consistency with human translators in English-to-Chinese translation is generally high, implying that models have some knowledge of the low frequency and the mainly negative context of Chinese passives, which helps them to make the right choice.

For future work, we intend to manually validate the annotation of the whole dataset, which will make the dataset serve better in a contrastive analysis of Chinese and English passives, and of passives in original and translational texts. As for model evaluation with the dataset, we intend to incorporate LLMs to better locate and categorize the corresponding translation of passive structures more precisely and automatically.

## 6. Limitations

Due to the uneven frequency of passive structures in Chinese and English, our dataset is unbalanced in terms of the number of passive sentences in the two languages, since we are collecting data from existing public parallel corpora to avoid copyright issues.

## 7. Acknowledgements

This work was supported by FairTransNLP-Language: Analysing toxicity and stereotypes in language for unbiased, fair and transparent systems (PID2021-124361OB-C33), funded by Ministerio de Ciencia, Innovación y Universidades, programa de I+D de Generación de Conocimiento (MICIU/AEI/10.13039/501100011033/FEDER, UE) and by Support grants to departments and university research units for the recruitment of pre-doctoral research staff in training (FI SDUR 2024).

## 8. Bibliographical References

## 9. Language Resource References